\documentclass{article}
\usepackage{spconf,amsmath,graphicx}
\usepackage{caption}
\usepackage{subcaption}
\usepackage{lineno,hyperref}
\usepackage{epstopdf}
\usepackage{graphicx}
\usepackage{url}
\usepackage{float}
\usepackage{amsfonts}

\title{Camera-Trap Images Segmentation using Multi-Layer Robust Principal Component Analysis}

\name{Jhony-Heriberto Giraldo-Zuluaga$^{\star}$, Alexander Gomez$^{\star}$, Augusto Salazar$^{\star}$, and Ang\'elica Diaz-Pulido$^{\dagger}$}
\address{$^{\star}$ Grupo de Investigaci\'on SISTEMIC, Facultad de Ingenier\'ia, Universidad de Antioquia UdeA,\\
Calle 70 No. 52-21, Medell\'in, Colombia \\
$^{\dagger}$ Instituto de Investigaci\'on de Recursos Biol\'ogicos Alexander von Humboldt,\\
Calle 28A No. 15-09, Bogot\'a D.C, Colombia}

\begin{document}
%
\maketitle
\begin{abstract}

Camera trapping is a technique to study wildlife using automatic triggered cameras. However, camera trapping collects a lot of false positives (images without animals), which must be segmented before the classification step. This paper presents a Multi-Layer Robust Principal Component Analysis (RPCA) for camera-trap images segmentation. Our Multi-Layer RPCA uses histogram equalization and Gaussian filter as pre-processing, texture and color descriptors as features, and morphological filters with active contour as post-processing. The experiments focus on computing the sparse and low-rank matrices with different amounts of camera-trap images. We tested the Multi-Layer RPCA in our camera-trap database. To our best knowledge, this paper is the first work proposing Multi-Layer RPCA and using it for camera-trap images segmentation.

\end{abstract}
\section{Introduction}
\label{sec:intro}

Studying and monitoring of mammals and birds species can be performed using non-invasive sampling techniques. These techniques allow us to observe animal species for conservation purposes, e.g. to estimate population sizes of endangered species. Camera trapping is a method to digitally capture wildlife images. This method facilitates the register of terrestrial vertebral species, e.g. cryptic species. Consequently, camera traps can generate large volumes of information in short periods of time. Thus, the contributions in camera trapping are important for better species conservation decisions.

Camera traps are devices to capture animal images in the wild. These devices consist of a digital camera and a motion detector. They are triggered when the motion sensor detects movement and dependent on the temperature of the source in relation to the environment temperature. Biologist can monitor wildlife with camera traps for detecting rare species, delineating species distributions, monitoring animal behavior, and measuring other biological rates \cite{o2010camera}. Camera traps generate large volumes of information, for example a camera trapping study can generate until 200000 images, where 1\% of the information is valuable \cite{diaz2011densidad}. As a consequence, biologists have to analyze thousands of photographs in a manual way. Nowadays, software solutions cannot handle the increment of the number of images in camera trapping \cite{fegraus2011data}. Accordingly, it is important to develop algorithms to assist the post-processing of camera-trap images.

Background subtraction techniques could help to segment animals from camera-trap images. There is a significant body research of background subtraction focused in video surveillance \cite{brutzer2011evaluation}. Nevertheless, there are not enough methods that can handle the complexity of natural dynamic scenes \cite{mahadevan2010spatiotemporal}. Camera-trap images segmentation is important for animal detection and classification. Camera-traps images usually have ripping water, moving shadows, swaying trees and leaves, sun spots, scene changes, among others. Consequently, the models used to segment those types of images should have robust feature extractors. There are some segmentation methods in the literature applied to camera-trap images segmentation. Reddy and Aravind proposed a method to segment tigers on camera-trap images, using texture and color features with active contours \cite{reddy2012segmentation}. They do not make an objective evaluation of their method. Ren et al. developed a method to segment images from dynamic scenes, including camera-trap images; the method uses Bag of Words (BoW), Histogram of Oriented Gradients (HOG), and graph cut energy minimization \cite{ren2013ensemble}. They do not show the results on camera-trap images. Zhang et al. developed a method to segment animals from video sequences, using camera-trap images; the method uses BoW, HOG, and graph cut energy minimization \cite{zhang2015coupled}. They obtained 0.8695 of average f-measure on their own camera-trap data set.

Robust Principal Component Analysis (RPCA) is a method derived from Principal Component Analysis. RPCA assumes that a data matrix can be decomposed in a low-rank and a sparse matrix. RPCA has newly seen significant activity in many areas of computer sciences, particularly in background subtraction. As a result, there are some algorithms to solve the RPCA problem \cite{liu2012active,lin2010augmented,goldfarb2013fast,aybat2011fast,wang2012probabilistic}. In this work, we proposed a Multi-Layer RPCA approach in order to segment animals from camera-trap images. Our method combines color and texture descriptors as feature extractor, and solve the RPCA problem with some state-of-the-art algorithms. To our knowledge, this paper is the first work in proposing a Multi-Layer RPCA approach and using it for camera-trap images segmentation.




The paper is organized as follows. Section \ref{sec:mateAndMeth} shows material and methods. Section \ref{sec:expFram} describes the experimental framework. Section \ref{sec:results} presents the experimental results and the discussion. Finally, Section \ref{sec:conclusions} shows the conclusions.

\section{Materials and Methods}
\label{sec:mateAndMeth}

This section shows a brief explanation of the algorithms and metrics used in this paper.

\subsection{Robust Principal Component Analysis}

An image can be decomposed in a low-rank and sparse matrix. Equation \ref{eqn:decomposition} shows the RPCA problem, where $M$ is the data matrix, $L_0$ is the low-rank matrix, and $S_0$ is the sparse matrix. The low-rank matrix is the background and the sparse matrix is the foreground in background subtraction.

\begin{equation}
    M=L_0+S_0
    \label{eqn:decomposition}
\end{equation}

The RPCA problem can be solved with the convex program Principal Component Pursuit (PCP). This program computes $L$ and $S$, taking the objective function in the Equation \ref{eqn:objectFunction}, where $||L||_*$ denotes the nuclear norm of the low-rank matrix, $||S||_1$ denotes the $l_1$-norm of the sparse matrix, and $\lambda$ is a regularizing parameter. There are some algorithms to perform PCP such as Accelerated Proximal Gradient (APG), and Augmented Lagrange multiplier (ALM) \cite{candes2011robust}.

\begin{equation}
    \begin{aligned}
    & \text{minimize}
    & & ||L||_* + \lambda ||S||_1 \\
    & \text{subject to}
    & & L+S=M
    \end{aligned}
    \label{eqn:objectFunction}
\end{equation}

\subsection{Multi-Layer Robust Principal Component Analysis}

Equation \ref{eqn:MMatrix} shows the data matrix $M$ in our Multi-Layer RPCA method, where $\beta \in [0,1]$ is a weight value indicating the contribution of the texture function to the overall data matrix. Function $f_t(x,y)$ denotes the texture descriptor extracted from each image, using the classic Local Binary Pattern (LBP) \cite{ojala2002multiresolution}. LBP describes the texture of an image using the neighborhood of each pixel. Function $f_c(x,y)$ denotes the color transformation for each image, converting to gray scale in this case. Our Multi-Layer RPCA computes the $L$ and $S$ matrices from the $M$ matrix in the Equation \ref{eqn:MMatrix}. Texture descriptors can work robustly into rich texture regions with light variation. However, they do not work in a efficient way on uniform regions such as water, the sky, and others. Color descriptors could overcome the texture descriptor limitation \cite{yao2007multi}. The Multi-Layer approach proposed in this paper was tested in camera traps for wildlife image segmentation.

\begin{equation}
M(x,y) = \beta f_t(x,y) + (1-\beta) f_c(x,y)
\label{eqn:MMatrix}
\end{equation}



\subsection{Evaluation Metrics}
\label{sec:evalMetrics}

The f-measure metric was chosen to evaluate the performance of the Multi-Layer RPCA. Equation \ref{eqn:fmeasure} shows the f-measure, where precision and recall are extracted from the confusion matrix. Precision is the proportion of predicted positives that are correctly real positives. In the same way, recall denotes the proportion of the real positives that are correctly predicted \cite{powers2011evaluation}. The confusion matrix is computed comparing the ground truth (GT) with the automatic segmented images.

\begin{equation}
\text{f-measure} = 2\frac{\text{precision}*\text{recall}}{\text{precision} + \text{recall}}
\label{eqn:fmeasure}
\end{equation}

\section{Experimental Framework}
\label{sec:expFram}

This section introduces the database used in this paper, the experiments executed, and the implementation details of our Multi-Layer RPCA.

\subsection{Database}
\label{subsec:database}

\begin{figure}
    \centering
    \begin{subfigure}[b]{0.15\textwidth}
		\includegraphics[width=\textwidth]{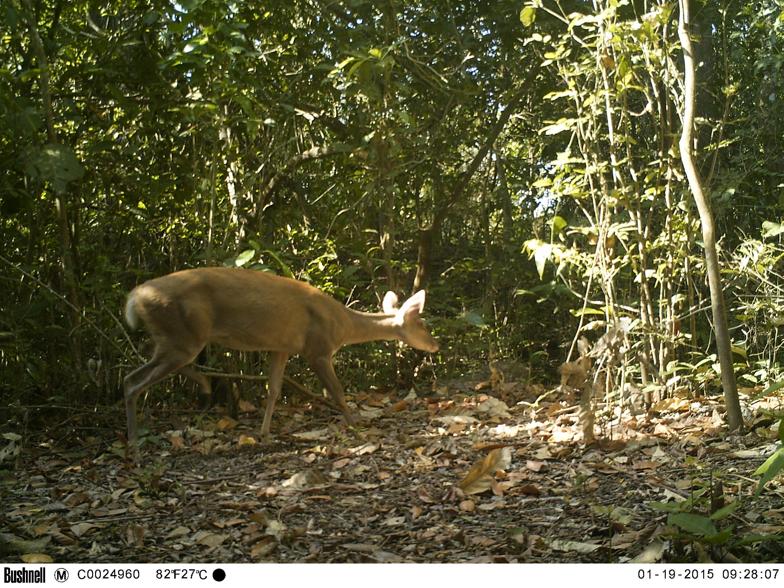}
        \caption{Original Image}
        \label{fig:GroundImage}
    \end{subfigure}
    \begin{subfigure}[b]{0.15\textwidth}
		\includegraphics[width=\textwidth]{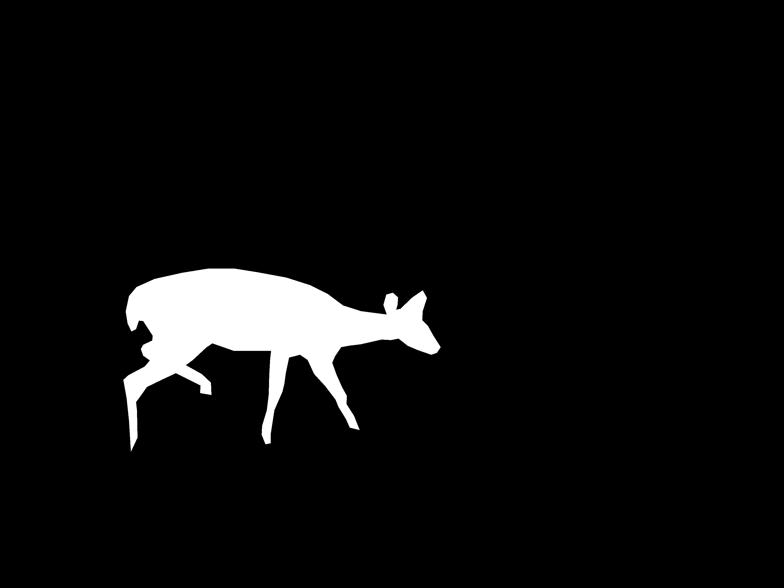}
        \caption{Ground truth}
        \label{fig:TruthImage}
    \end{subfigure}
    \caption{Ground truth example for the evaluation process, (a) original image, (b) ground truth or manual segmented image.}
    \label{fig:GTImages}
\end{figure}

The Alexander von Humboldt Institute realizes samplings with camera traps in different regions of the Colombian forest. We select 25 cameras from 8 regions, where each camera has a relative unalterable environment. Each camera was placed in its site between 1 to 3 months. We extract 30 days and 30 nights of images from those cameras, in daytime color and nighttime infrared formats respectively. The database consists of 1065 GT images from the 30 days and 30 nights. The images have a spatial resolution of 3264x2448 pixels. The length of each day or night data set varies from 9 to 72 images, depending on the animal activity that day or night. Figure \ref{fig:GTImages} shows an example of the GT images.

\subsection{Experiments}

\begin{figure}
    \centering
    \includegraphics[width=0.38\textwidth]{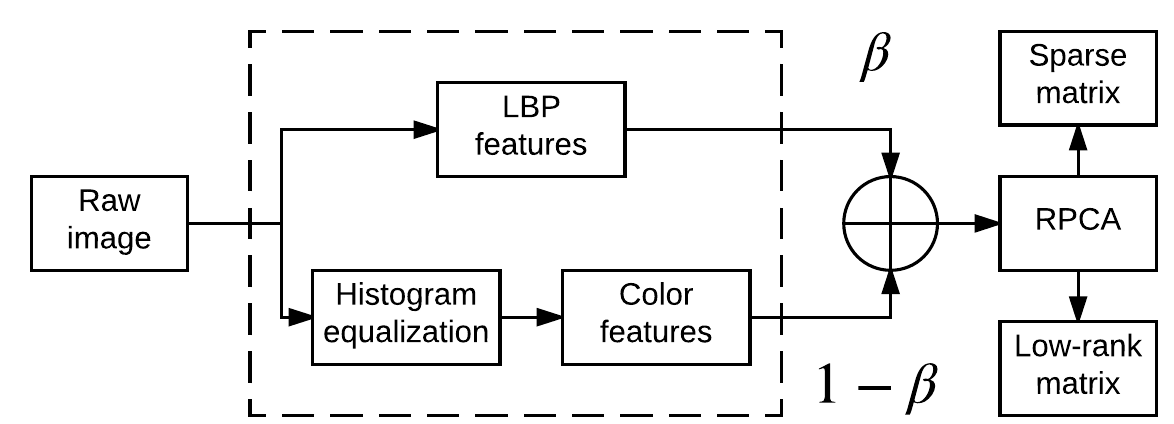}
    \caption{Block diagram of the pre-processing methods used in the Experiment 1, 3, and 4.}
    \label{fig:preprocess1}
\end{figure}

\begin{figure}
    \centering
    \includegraphics[width=0.38\textwidth]{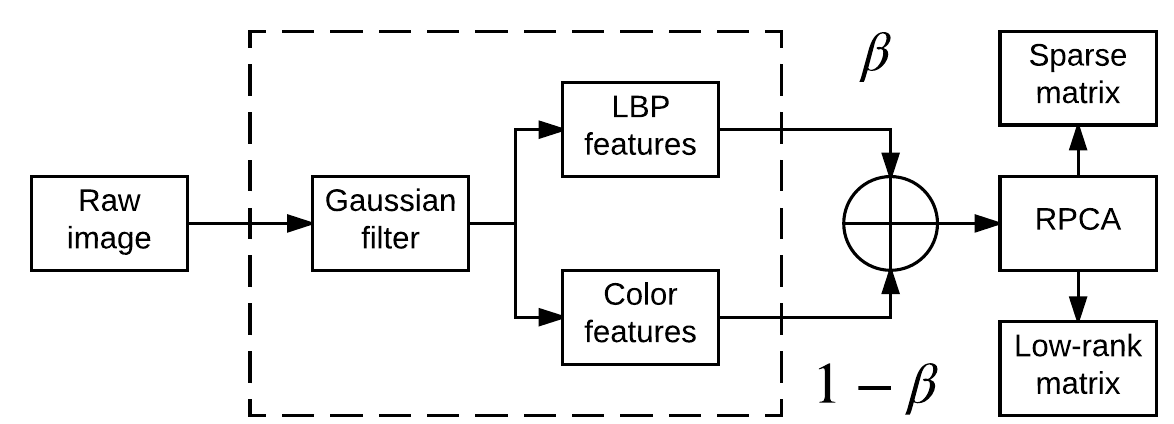}
    \caption{Block diagram of the pre-processing methods used in the Experiment 2 and 3.}
    \label{fig:preprocess2}
\end{figure}

The experiments computed the background models with different conditions and amount of images, observing the robustness of the Multi-Layer RPCA and the influence of pre-processing in the results. All experiments performed our method with $\beta=[0,0.05,0.1,0.15,\dots,1]$. Accordingly, Experiment 1 uses histogram equalization as pre-processing in the color transformed image, Figure \ref{fig:preprocess1} shows the pre-processing for each raw image. The background model is computed with entire days e.g. we take all images of day 1 and solve the RPCA problem in the Experiment 1. Experiment 2 computes the background model with entire nights; Figure \ref{fig:preprocess2} shows the pre-processing for each raw image. Experiment 3 takes entire days and nights e.g. we take all images of day 1 and night 1 to solve the RPCA problem. Experiment 3 uses two pre-processes, daytime images uses the pre-process in Figure \ref{fig:preprocess1} and nighttime images uses the pre-process in Figure \ref{fig:preprocess2}. Experiment 4 takes entire days and nights such as the Experiment 3, but it only uses the pre-processing in Figure \ref{fig:preprocess1} for all images.


We tested 9 algorithms to solve the RPCA problem in this paper. Active Subspace RPCA (AS-RPCA) \cite{liu2012active}; Exact ALM (EALM), Inexact ALM (IALM), Partial APG (APG-PARTIAL) and APG \cite{lin2010augmented}; Lagrangian Alternating Direction Method (LSADM) \cite{goldfarb2013fast}; Non-Smooth Augmented Lagrangian v1 (NSA1) and Non-Smooth Augmented Lagrangian v2 (NSA2) \cite{aybat2011fast}; Probabilistic Robust Matrix Factorization (PRMF) \cite{wang2012probabilistic}.

\begin{figure}
    \centering
    \includegraphics[width=0.22\textwidth]{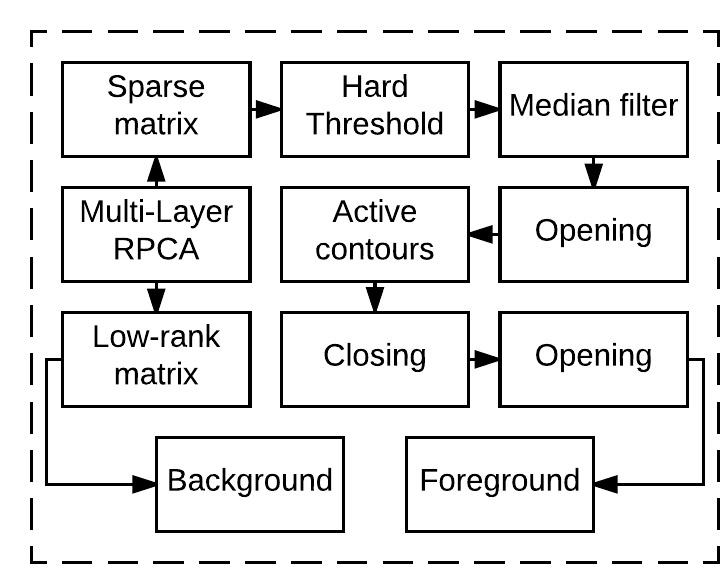}
    \caption{Block diagram of the postprocessing.}
    \label{fig:postprocess}
\end{figure}

The foreground was obtained applying a post-process to the sparse matrix. The post-processing was the same for all experiments. This stage includes a hard threshold, morphological filters, and an active contours with a negative contraction bias \cite{caselles1997geodesic}. Figure \ref{fig:postprocess} shows the post-processing used. Finally, The f-measure was computed comparing each GT with each foreground. The average f-measure was computed as the mean of all f-measures. The results are displayed as a plot of the average f-measure vs $\beta$.

\subsection{Implementation Details}

The RPCA algorithms were computed using the Sobral et al. library \cite{lrslibrary2015}. The rest of the source code was developed using the image processing toolbox of Matlab.

\section{Results}
\label{sec:results}

This section shows the results and discussions of the experiments introduced in the Section \ref{sec:expFram}. These results use the metrics explained in the Section \ref{sec:evalMetrics}.

\begin{figure*}
    \centering
    \begin{subfigure}[b]{0.36\textwidth}
		\includegraphics[width=\textwidth]{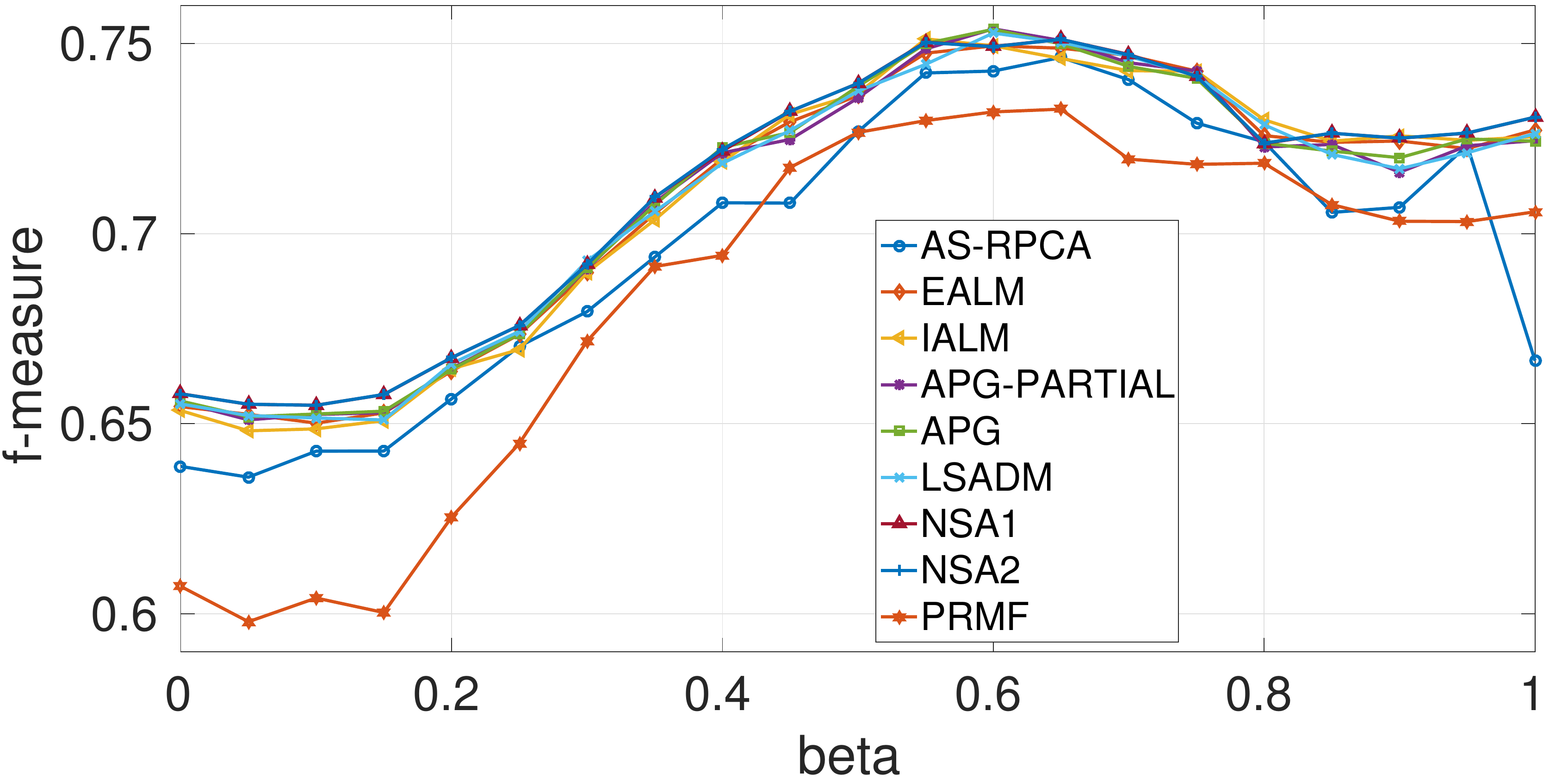}
        \caption{Average f-measure vs $\beta$ for the experiment 1}
        \label{fig:ResultExperiment1}
    \end{subfigure}
    \begin{subfigure}[b]{0.36\textwidth}
		\includegraphics[width=\textwidth]{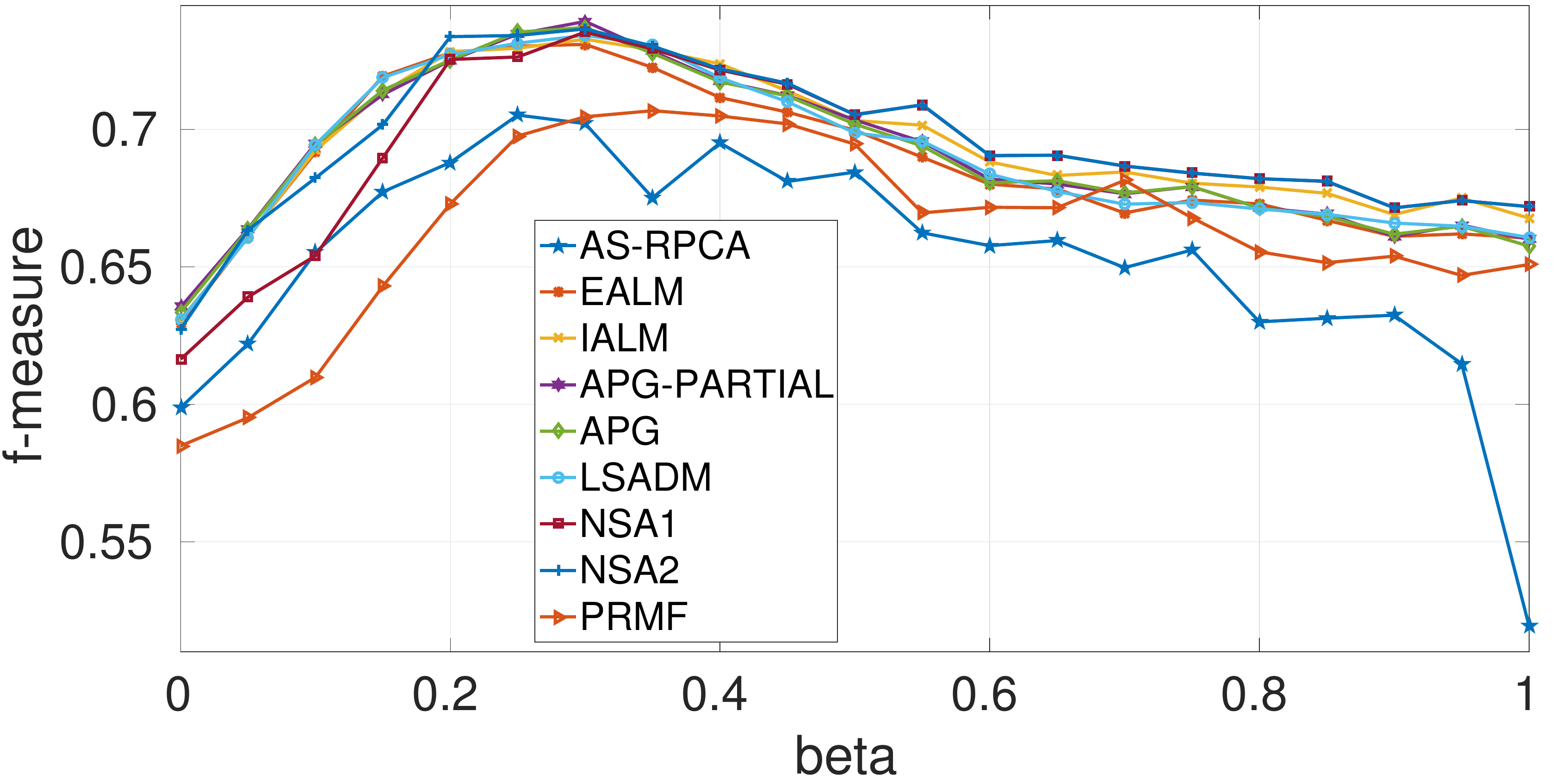}
        \caption{Average f-measure vs $\beta$ for the experiment 2}
        \label{fig:ResultExperiment2}
    \end{subfigure}
    \begin{subfigure}[b]{0.36\textwidth}
		\includegraphics[width=\textwidth]{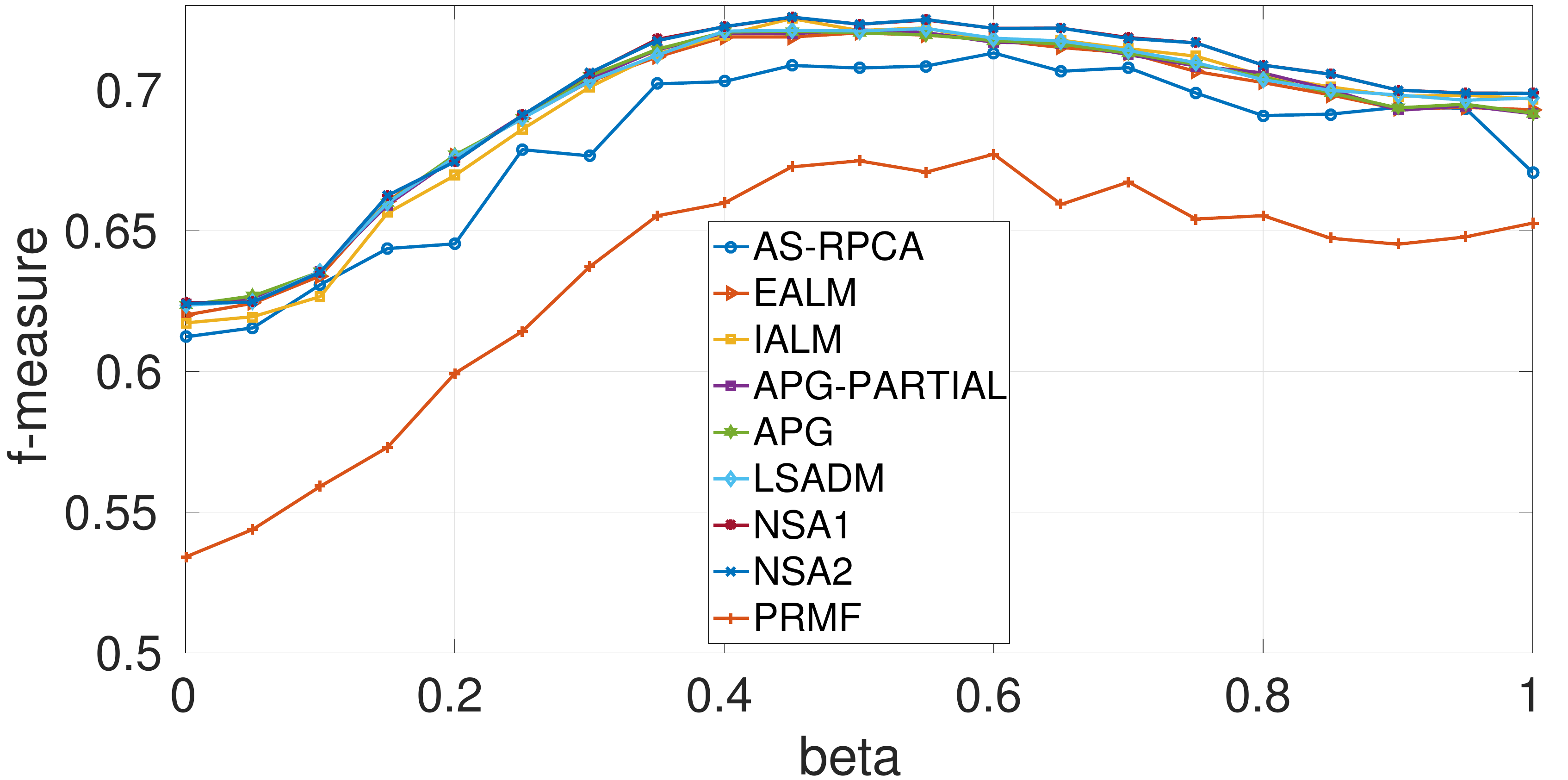}
        \caption{Average f-measure vs $\beta$ for the experiment 3}
        \label{fig:ResultExperiment3}
    \end{subfigure}
    \begin{subfigure}[b]{0.36\textwidth}
		\includegraphics[width=\textwidth]{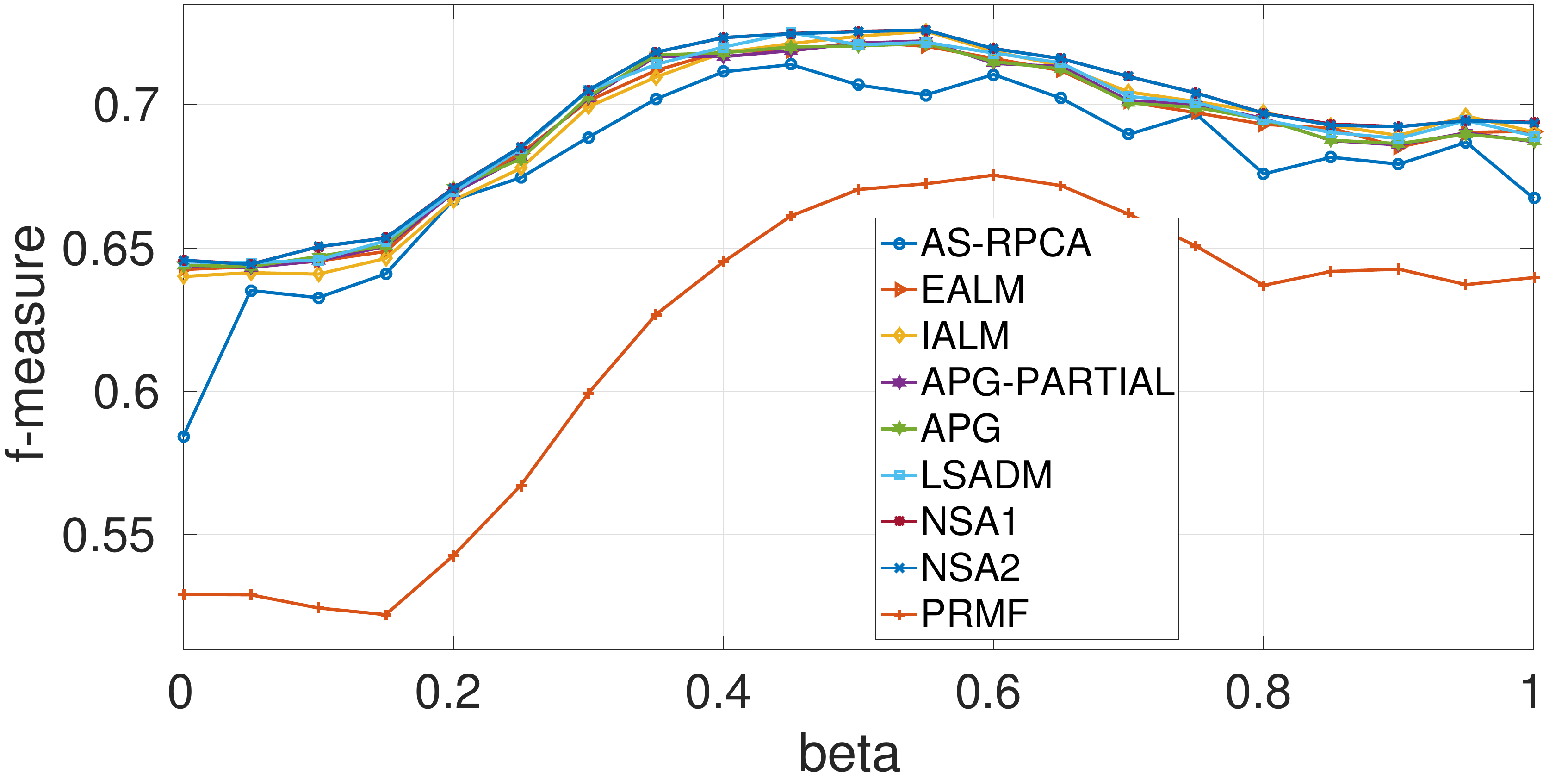}
        \caption{Average f-measure vs $\beta$ for the experiment 4}
        \label{fig:ResultExperiment4}
    \end{subfigure}
    \caption{Results of the proposed experiments, (a) average f-measure vs $\beta$ per days, (b) average f-measure vs $\beta$ per nights, (c) average f-measure vs $\beta$ per days and nights with two different pre-processes, (d) average f-measure vs $\beta$ per days and nights with one pre-process.}
    \label{fig:resultsExperiments}
\end{figure*}

Figures \ref{fig:ResultExperiment1} and \ref{fig:ResultExperiment2} show the average f-measure vs $\beta$ of the Experiments 1 and 2 for all RPCA algorithms chosen. Table \ref{tbl:summary} shows the summary of the best results for each experiment. APG-PARTIAL was the best algorithm in the Experiments 1 and 2. Daytime images have rich texture regions. In contrast, nighttime images have uniform color. Texture representations are more important on daytime images in the Experiment 1 due to $\beta=0.6$. On the contrary, color descriptors are more important on nighttime images in the Experiment 2 due to $\beta=0.3$. Those results show the importance of combining the color and texture descriptors. Figure \ref{fig:resultsExperiments} shows the performances of the RPCA normal algorithms when $\beta=0$. Thus, our Multi-Layer RPCA outperforms the RPCA normal methods.


Figures \ref{fig:ResultExperiment3} and \ref{fig:ResultExperiment4} show the average f-measure vs $\beta$ of the Experiments 3 and 4. Table \ref{tbl:summary} shows that dividing the pre-processing per daytime or nighttime in the Experiment 3 does not make a big difference in the results, but it increases the fine-tuning parameters. Table \ref{tbl:summary} shows that NSA2 was the best algorithm in the Experiments 3 and 4, contrary to the Experiments 1 and 2 where APG-PARTIAL was the best. NSA2 algorithm is a better choice than other RPCA algorithms, if we cannot differentiate between daytime and nighttime images, or if it is difficult to do so. On the other hand, APG-PARTIAL is better, if we have information about the infrared activation.



\begin{figure}
    \centering
    \begin{subfigure}[b]{0.12\textwidth}
		\includegraphics[width=\textwidth]{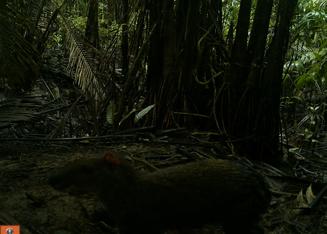}
        \caption{Ground}
        \label{fig:ground1}
    \end{subfigure}
    \begin{subfigure}[b]{0.12\textwidth}
		\includegraphics[width=\textwidth]{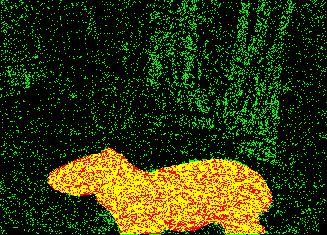}
        \caption{Sparse}
        \label{fig:sparse1}
    \end{subfigure}
    \begin{subfigure}[b]{0.12\textwidth}
		\includegraphics[width=\textwidth]{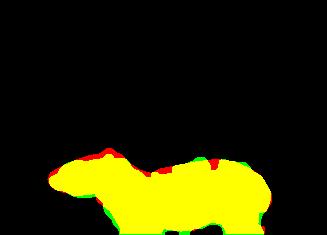}
        \caption{Foreground}
        \label{fig:foreground1}
    \end{subfigure}
    \begin{subfigure}[b]{0.12\textwidth}
		\includegraphics[width=\textwidth]{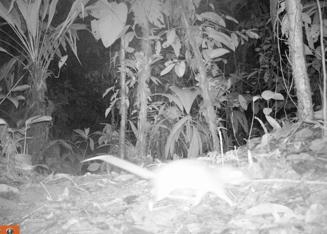}
        \caption{Ground}
        \label{fig:ground2}
    \end{subfigure}
    \begin{subfigure}[b]{0.12\textwidth}
		\includegraphics[width=\textwidth]{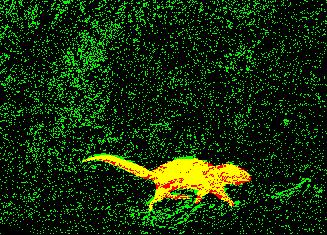}
        \caption{Sparse}
        \label{fig:sparse2}
    \end{subfigure}
    \begin{subfigure}[b]{0.12\textwidth}
		\includegraphics[width=\textwidth]{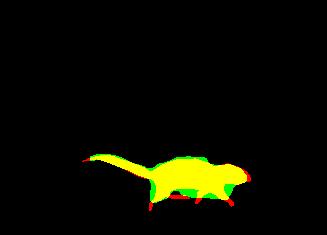}
        \caption{Foreground}
        \label{fig:foreground2}
    \end{subfigure}
    \caption{Visual results of the Multi-Layer RPCA, (a) original daytime image, (b) sparse matrix after the hard threshold with APG-PARTIAL and $\beta=0.6$, (c) foreground, (d) original nighttime image, (e) sparse matrix after the hard threshold with APG-PARTIAL and $\beta=0.3$, (f) foreground.}
    \label{fig:visualResults}
\end{figure}

Figure \ref{fig:visualResults} shows two visual results of the Multi-Layer RPCA. Figure \ref{fig:ground1} shows a daytime image without any pre-processing. Figure \ref{fig:ground2} shows an original nighttime image. Figures \ref{fig:sparse1} and \ref{fig:sparse2} show the sparse matrix after the hard threshold. Figures \ref{fig:foreground1} and \ref{fig:foreground2} show the foreground image. These color results are made with the GT images. Yellow-colored regions mean pixels that are on the GT and the automatic segmented images. Red and green regions are visual representations of the under and over segmentation.


\begin{table}
\centering
\caption{$\beta$ values and algorithms for the best performances of each experiment.}
\label{tbl:summary}
\begin{tabular}{cccc}
\hline
\textbf{Experiment} & \textbf{Algorithm} & $\boldsymbol{\beta}$ & \textbf{Avg f-measure} \\ \hline
Experiment 1 & APG-PARTIAL & 0.6  & 0.7539 \\ 
Experiment 2 & APG-PARTIAL & 0.3  & 0.7393 \\ 
Experiment 3 & NSA2        & 0.45 & 0.7259 \\ 
Experiment 4 & NSA2        & 0.55 & 0.7261 \\ \hline
\end{tabular}
\end{table}

\section{Conclusions}
\label{sec:conclusions}

We proposed a Multi-Layer RPCA for camera-trap image segmentation, using texture and color descriptors. The proposed algorithm is composed of pre-processing, RPCA algorithm, and post-processing. The pre-processing uses histogram equalization, Gaussian filtering, or a combination of both. The RPCA algorithm computes the sparse and low-rank matrices for background subtraction. The post-processing computes morphological filters and an active contours with a negative contraction bias. We proved the Multi-Layer RPCA algorithm in a camera-trap images database from the Colombian forest. The database was manually segmented to extract the f-measure of each automatic segmented image. We reach 0.7539 and 0.7393 of average f-measure in daytime and nighttime images respectively. The average f-measure was computed with all GT images. To our best knowledge, this paper is the first work in proposing Multi-Layer RPCA and using it for camera-trap images segmentation.\\
\\
\textbf{Acknowledgment.} This work was supported by the Colombian National Fund for Science, Technology and Innovation, Francisco Jos\'e de Caldas - COLCIENCIAS (Colombia). Project No. 111571451061.

\bibliographystyle{ieeetr}
\bibliography{bibfile}

\end{document}